\documentclass[letterpaper]{article} 
\usepackage{aaai25}  
\usepackage{times}  
\usepackage{helvet}  
\usepackage{courier}  
\usepackage[hyphens]{url}  
\usepackage{subcaption}
\usepackage{amsmath}
\usepackage{amsfonts}
\usepackage{xcolor}
\usepackage{graphicx} 
\usepackage{natbib}  
\usepackage{caption} 
\usepackage{algorithm}
\usepackage{algorithmic}

\usepackage{newfloat}
\usepackage{listings}
\DeclareCaptionStyle{ruled}{labelfont=normalfont,labelsep=colon,strut=off} 
\floatstyle{ruled}
\newfloat{listing}{tb}{lst}{}
\floatname{listing}{Listing}
\title{
Spatial Visibility and Temporal Dynamics:\\ Revolutionizing Field of View Prediction in Adaptive Point Cloud Video Streaming
}
\author {
    Chen Li,
    Tongyu Zong,
    Yueyu Hu,
    Yao Wang,
    Yong Liu
}
\affiliations {
    Department of Electrical and Computer Engineering\\
    Tandon School of Engineering\\
    New York University\\
    \{chen.lee, tz1178, yh3986, yw523, yongliu\}@nyu.edu
}
\usepackage{bibentry}

\makeatletter
\def\@copyrightspace{\vspace{0pt}}
\makeatother

\begin{document}

\maketitle

\begin{abstract}

Field-of-View (FoV) adaptive streaming significantly reduces bandwidth requirement of immersive point cloud   video (PCV) by only transmitting visible points in a viewer's FoV. The traditional approaches often focus on trajectory-based 6 degree-of-freedom (6DoF) FoV predictions. The predicted FoV is then used to calculate point visibility. Such approaches do not explicitly consider video content's impact on viewer attention, and the conversion from FoV to point visibility is often error-prone and time-consuming. We reformulate the PCV FoV prediction problem from the cell visibility perspective, allowing for precise decision-making regarding the transmission of 3D data at the cell level based on the predicted visibility distribution. We develop a novel spatial visibility and object-aware graph model that leverages the historical 3D visibility data and incorporates spatial perception, neighboring cell correlation, and occlusion information to predict the cell visibility in the future. 
Our model significantly improves the long-term cell  visibility prediction, reducing the prediction MSE loss by up to 50\% compared to the state-of-the-art models while maintaining real-time performance (more than 30fps) for point cloud videos with over 1 million points.
\end{abstract}

%

\section{Introduction}

\begin{figure}[ht]
    \centering
    \includegraphics[width=0.9\linewidth]{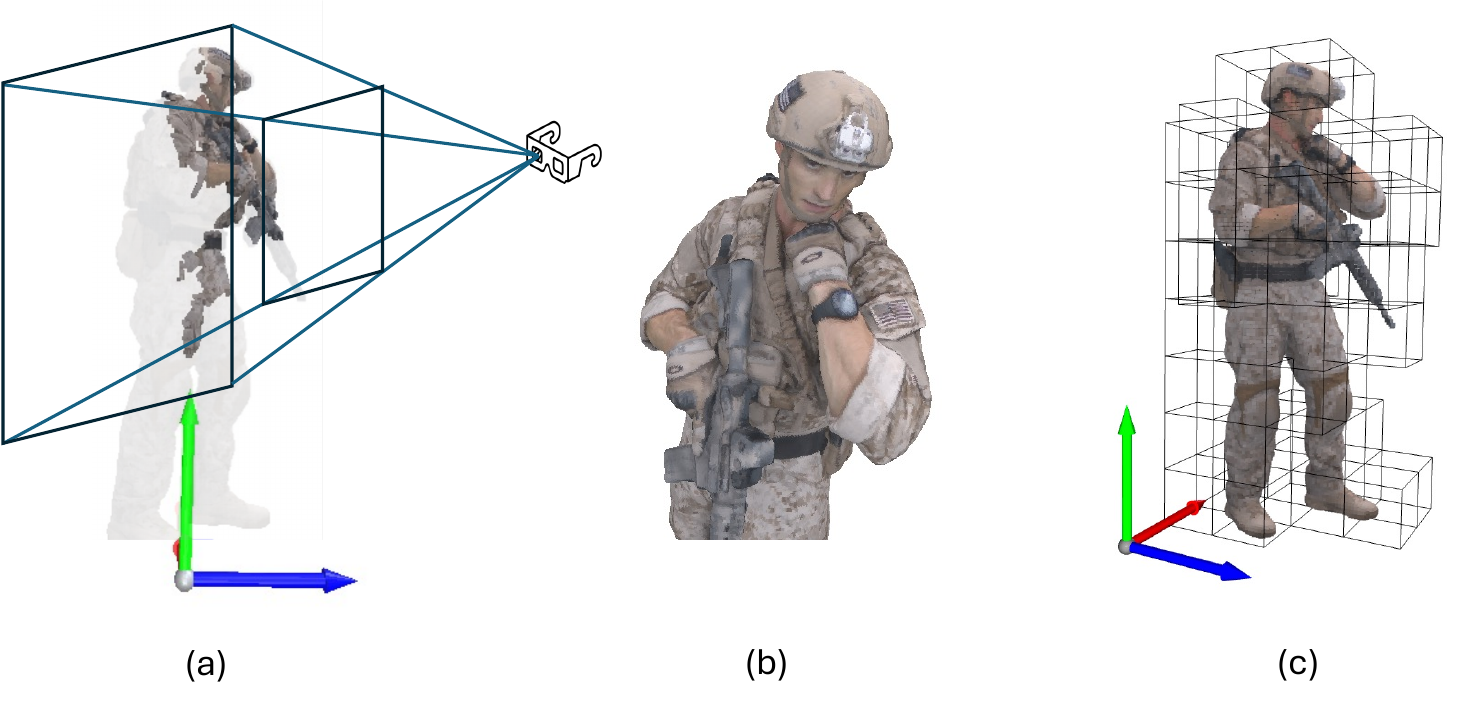}
    \caption{6DoF FoV Demonstration. As in Fig.(a), the content within the pyramid which is bounded with far plane and near plane is inside the viewport and any content outside pyramid will not be seen by viewer. Furthermore, even though some points are inside the viewport, they are still not visible if occluded by other points. So if there is a viewer watching the point cloud content from side as Fig. 1(a) shows (the actual view is shown in Fig.1(b)), only the highlighted part in Fig. 1(a) would be the visible. If the point cloud is divided into  3D cells like in Fig. 1(c), only the cells covering the visible points need to be transmitted.}
    \label{fig:fov_demo}
\end{figure}

AR/VR applications are gaining popularity rapidly. Streaming high-quality immersive videos, such as 360-degree videos and point cloud videos, to viewers is one of the most critical components for the wide adoption of AR/VR. Immersive videos require significantly higher bandwidth than the traditional 2D planar videos. For example, a point cloud video consisting of 300k to 1M points requires streaming bandwidth from 1.08 Gbps to 3.6 Gbps~\cite{groot2020}. A promising solution is FoV adaptive streaming that only streams video content within a viewer's current viewport. For example, for 360-degree video, if the viewport is 120 degree (horizontal) by 90 degree (vertical), only 1/6 video data fall into a viewport, resulting into 6-fold bandwidth reduction. For point cloud video, not all points falling into a viewport are visible, due to points occlusion in 3D space.  One can save even more bandwidth by not just omitting the points outside the viewport, but also removing the hidden points within the viewport. As illustrated in Fig.~\ref{fig:fov_demo}, the number of the actual visible points in Fig.~\ref{fig:fov_demo}(b) is less than 150k after removing the points outside the viewport and the occluded points. Even though for a single object point cloud frame, compared with the full content with over 1M points, we can already save up to 7 times of bandwidth potentially.

One key challenge for FoV adaptive streaming is viewer FoV prediction. In immersive video streaming, a viewer can freely change her viewpoint $(X, Y, Z)$,  as well as her view angle {\it (yaw, pitch, roll)}, resulting in a total of 6DoF. To prefetch visible part into the client buffer, FoV adaptive streaming has to predict the viewer's future FoV to determine which portion of video content will fall into her viewport. For on-demand streaming, a relative long streaming buffer, e.g., 2 to 5  seconds, is preferred to provide sufficient margin for smooth video streaming and video processing. Consequently, accurate long-term viewport prediction is essential for FoV adaptive streaming, and has become an active research topic for both 360 degree video~\cite{longterm360Dong} and point cloud video~\cite{han2020vivo,liu2023cav3}. For point cloud video, based on the predicted viewport, one needs to further conduct Hidden Point Removal (HPR) to filter out points that deemed not visible to the viewer. A PCV frame is typically divided into cells, and each cell is independently coded and transmitted, as shown in Fig.~\ref{fig:fov_demo}(c). Given the point visibility, the system calculates the visibility of a cell as the number of visible points in the cell, and determine cell streaming rate based on its  visibility.  

Naturally, most of the existing FoV-adaptive PCV streaming studies~\cite{hou2020motion,liu2023cav3,han2020vivo,zong2023progressive} predict visible points in two steps: 1) predict the viewer's future viewport based on her past viewport trajectory; 2) conduct HPR using the predicted viewport. This approach has several drawbacks: 1) trajectory-based viewport prediction does not explicitly consider the impact of video content on the viewer's attention; 2) small errors in 6DoF viewport coordinates prediction may lead to large errors in visible points prediction; and 3) HPR at high point density is time consuming.

This motivates us to rethink FoV prediction for PCV streaming:    %
{\it Can we directly predict cell visibility based on the viewer's viewport trajectory, cell visibility history, and spatial features of objects to be viewed in PCV?}  There are several potential advantages of this direct approach: 1) by directly predicting cell visibility based on the cell visibility history (we assume that the viewer's past viewport is fedback to the sender so that the sender can produce the visibility history data), we avoid the potential error amplification in the process of mapping 6DoF viewport to cell visibility; 2) using the spatial features of PCV objects to be viewed (for which we also have the ground-truth in on-demand streaming) as an input, we explicitly take into account the impact of PCV objects and their movements on the viewer's attention; 3) PCV object movements and viewer viewpoint movements are both continuous in time and space. As a result, point occlusion and visibility vary continuously in time and space. For example, the increasing visibility of one part will give a strong hint that its neighbor could become more visible in the future. A well-trained spatial-temporal machine learning model can leverage such continuities for accurate point/cell visibility prediction. Towards realizing these potential gains, we make the following contributions in this paper: 
\begin{itemize}
    \item We design a novel PCV FoV prediction framework that directly predicts the cell visibility in a future frame to a viewer based on the viewer's past viewport trajectory and the point cloud spatial features to be viewed. 
    \item We develop a spatial-temporal graph model which can capture the spatial and temporal correlations of cell visibility for accurate prediction.
    \item Through case studies, we demonstrate that our model can improve the long-term cell visibility prediction accuracy by up to 50\% on the real point cloud video and viewport trajectory datasets in real-time.
\end{itemize}



\section{Related Work}
Immersive video transmission is considered to be bandwidth consuming, and predicting users’ preferences can save bandwidth and reduce latency, like pre-caching user-preferred contents in traditional 2D video~\cite{li2023predictive} and predicting user's future viewport in 360-degree videos~\cite{wu2020spherical}. There have been many work developing FoV prediction techniques to reduce bandwidth consumption while maximally maintaining the user experience~\cite{ban2018cub360,qian2016optimizing,fan2017fixation}. To achieve longer prediction horizon with higher accuracy to facilitate high quality streaming, \cite{li2019very} proposed several approaches for predicting future FoV from the past viewport trajectory using variations of Long Short-Term Memory (LSTM) models, with or without knowledge of other users's FoV for watching the same video. It also develop approaches for predicting the tile visibility using 2D spatial and temporal ConvLSTM  model.

However, 360-degree videos assume all pixels are potentially visible to the user without occlusion, whereas in 6DoF point cloud videos, cells in the front can block those in the back. It makes prediction of visibility more challenging. To tackle this problem, 
in~\cite{hou2020motion}, the authors develop a user motion and a head orientation prediction model using LSTM and MLP to together predict the 6DoF FoV. This method is capable of predicting the FoV for the next frame accurately so the streaming system can render the next frame in advance to improve the Quality of Experience (QoE).
Despite that, the narrow prediction horizon (\textit{i.e.} 11 ms for the next frame only) significantly limits the available optimization time span given to the streaming system to adjust the strategy, hence it makes the system vulnerable to network fluctuation. The work Vivo~\cite{han2020vivo} extend the prediction horizon to 200 ms, and Cav3 Gaze~\cite{liu2023cav3} further reduce the orientation coordinate degree to further improve the prediction accuracy. However, the prediction horizon is still limited for more sophisticated streaming strategies.
In this work, we manage to expand the prediction horizon to as long as 5000 ms, which greatly improve the resilience of the adaptive streaming system by allowing more processing time and more robustness to network fluctuation.

\section{Background and Problem Formulation}

\subsection{Point Cloud Video}
\begin{figure*}[ht]
    \centering
    \includegraphics[scale=0.65]{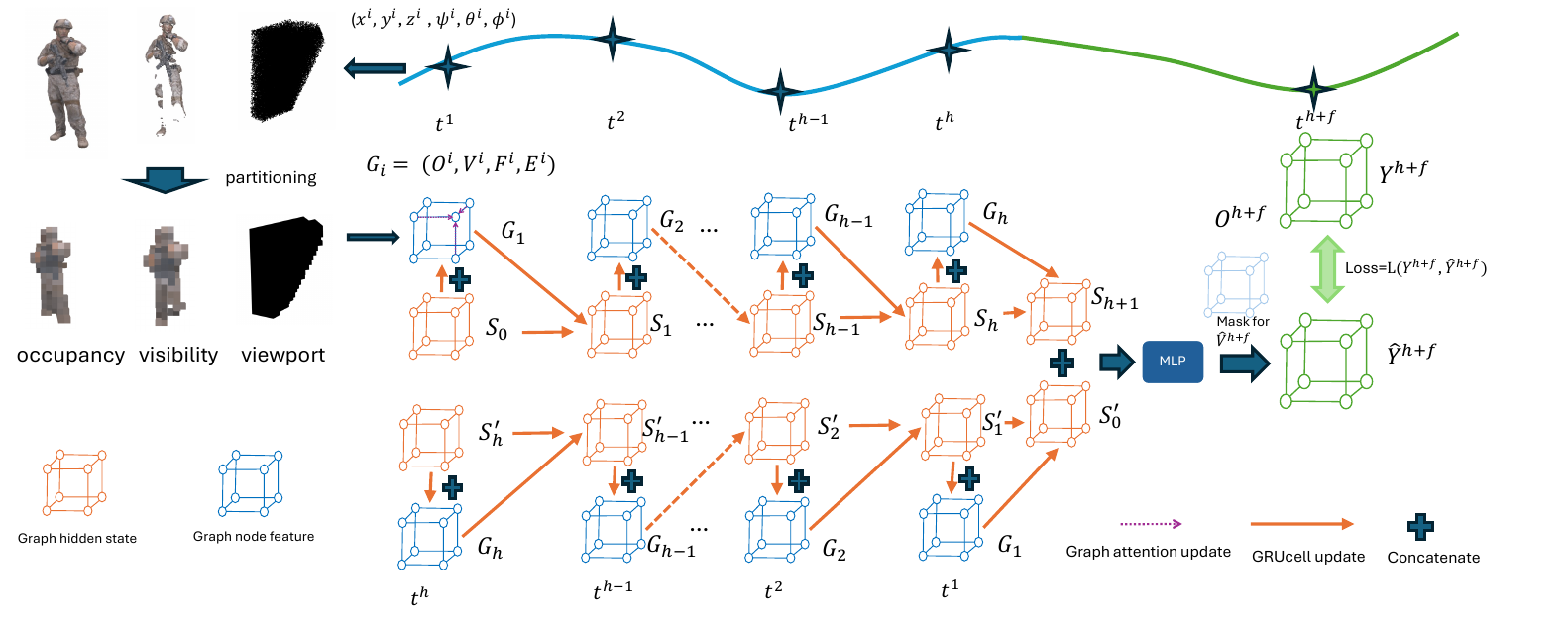}
    \caption{Overview of our cell visibility prediction system. 
    The curve on the top is the viewer's 6DoF viewport trajectory, illustrating one of the the 6DoF coordinates $ (x,y,z,\psi,\theta, \phi) $ at each frame time $t$  for a point cloud sequence set $\mathcal{P}^{h+f}$.
    We partition the whole space into 3D cells and model the cells into 3D grid-like graph. For each frame $t$, based on the 6DoF coordinates and $P^i$, we can calculate, for each cell $i$,  the total number of points $O_i$, visible points $V_i$, and percent of cell volume within the viewport (called the cell-based viewport feature) $F_i$. 
    We use  $G_i=[O_i, F_i, V_i, E_i]$ to represent the node feature for each frame, where $E_i$ is other features like the coordinates of each cell. We use a  temporal Bidirectional GRU model and a spatial transformer-based graph model to capture patterns in viewer attention and cell visibility. GRU model captures each node's temporal pattern over time, which is encoded into the hidden state $S_h$. In the graph model, each node aggregates its neighbor's information and hidden state from GRU, as the dash line shows. To simplify the graph, we only show the graph attention updating on the $G_1$. After we got the $S_{h+1}$ and $S^{'}_{0}$ from bi-directional GRU, a MLP modal will predict the cell visibility at the target time stamp $t^{h+f}$. Before the final output, the $O^{h+f}$ will be applied as mask to get the predicted visibility $\hat{V}^{h+f}$, since in the streaming system, the server has the point cloud at frame $h+f$. 
    We can optionally predict the overlap ratios between viewport and cells, $\hat{F}^{h+f}$, as well (which essentially predict the viewport).  
    The $\hat{Y}^{h+f}$ is the prediction output, indicating either  $\hat{V}^{h+f}$ or $\hat{F}^{h+f}$.
    }
    \label{fig:system}
\end{figure*}
Point cloud video (PCV) is a new immersive video format captured by multiple depth cameras.  PCV consists of a sequence of frames, each of which is a cloud of  
points described by their 3D coordinates and colors. It can be rendered on 2D displays or VR goggles based on the user's view point, allowing users to explore scenes from any angle and depth with 6 degrees of 
freedom. It provides viewers with the most immersive experience among other video technology so far. For streaming applications, each PCV frame is typically partitioned into small 3D cells, which can be encoded independently at multiple quality levels~\cite{wang2021qoe}, as shown in Fig.~\ref{fig:fov_demo}(c). For a given viewer viewport, only a small subset of cells 
are visible. If we know exactly which cells are visible, we just need to  stream those visible cells at the highest quality allowed by the bandwidth. To deal with the 
unavoidable cell visibility prediction errors, we can also stream some cells with low visibility at low quality similar to the 360-degree video streaming strategy~\cite{sun2019two}. The final PCV quality perceived by a viewer and PCV streaming overhead is largely determined by the accuracy of cell visibility prediction.  
\begin{figure}[ht]
    \centering
    \includegraphics[width=0.7\linewidth]{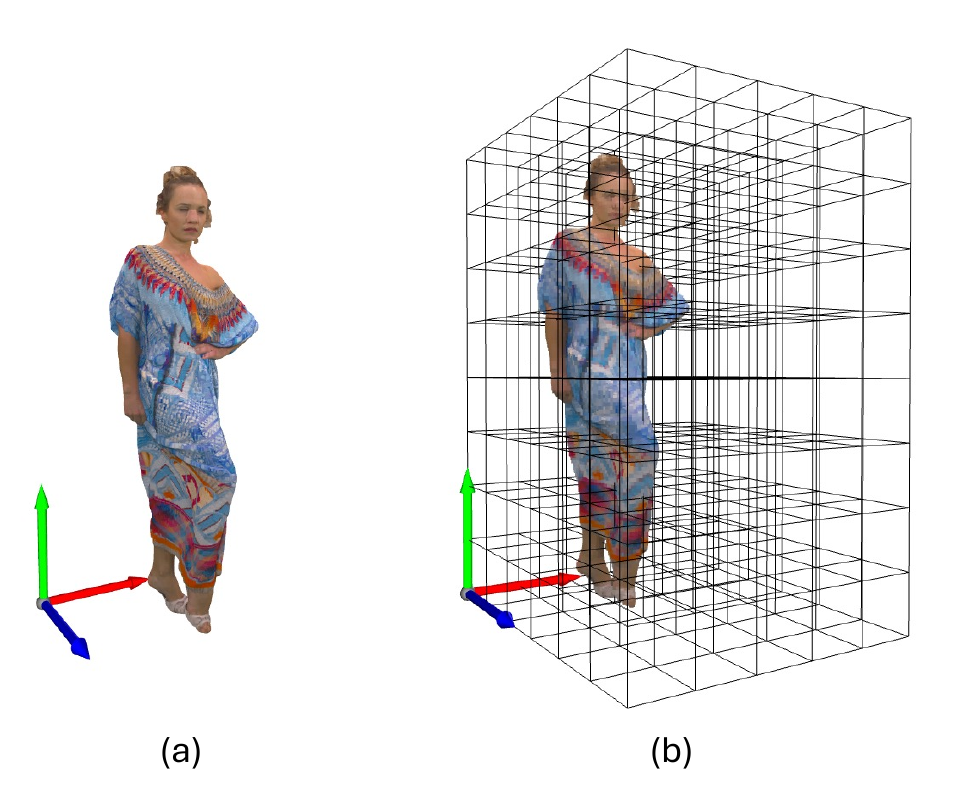}
    \caption{(a) is an example frame of a point cloud video.  F(b) is the graph we build for the full scene video and each grid is a node in graph. Since the object in the video is moving, we voxelize the whole space into grids.}
    \label{fig:3dcell}
\end{figure}

\subsection{Problem Formulation}
\begin{table}[ht]
  \centering
  \begin{tabular}{|c|c|}
    \hline
    \textbf{Symbol} & \textbf{Description} \\ \hline
    $T_{\tau}$ & 6-DoF coordinates trajectory at time $\tau$ \\ \hline
    $\mathcal{T}_h$ & history trajectory \\ \hline
    $C$ & 3D cell set \\ \hline
    $O^{\tau}$ & graph occupancy features at time $\tau$ \\ \hline
    $V^{\tau}$ & graph visibility features at time $\tau$ \\ \hline
    $F^{\tau}$ & graph viewport features at time $\tau$ \\ \hline
    $E^{\tau}$ & graph other features at time $\tau$ \\ \hline
    $G^{\tau}$ & all graph raw features at time $\tau$ \\ \hline
    $S_{\tau}$ & hidden states at time $\tau$ \\ \hline
    $o^{\tau}_i$ & occupancy feature of cell $i$ at time $\tau$ \\ \hline
    $v^{\tau}_i$ & visibility feature of cell $i$ at time $\tau$ \\ \hline
    $f^{\tau}_i$ & viewport feature of cell $i$ at time $\tau$ \\ \hline
    $e^{\tau}_i$ & other features of cell $i$ at time $\tau$ \\ \hline
  \end{tabular}
  \caption{Notation table}
  \label{tab:notation}
\end{table}
We denote the viewer's history viewport trajectory by
$$
\mathcal{T}_h: \{T^1, ..., T^h\}
$$
where $T^{\tau} = (x^{\tau},y^{\tau},z^{\tau},\psi^{\tau},\theta^{\tau},\phi^{\tau})$ is the 6-DoF coordinates at time $\tau$, and the corresponding PCV frame sequence by  $\mathcal{P}_h = \{P^1, ..., P^h \}$. 
Each frame $P^\tau$ is partitioned into a set $C$ of cells, each cell contains a set of points. The visibility of a cell $i$ at time $\tau$ is defined as the number of visible points in that cell, denoted by as $v_{i}^\tau$. The cell visibility vector for frame $\tau$ is $V^\tau = \{ v_{i}^\tau, i \in C\}$. We assume that the  client has a frame buffer of length $f$, so that when the client is displaying frame $h$, the server needs to send frame $h+f$.  Our goal is to predict the cell visibility for a future frame $h+f$ from $\mathcal{T}_h,\mathcal{P}_{h},P^{h+f}$ using a learnt function $\mathcal{F}$:
\[
V^{h+f}=\mathcal{F}(\mathcal{T}_h,\mathcal{P}_{h},P^{h+f}). 
\]

\section{Methodology}
\label{sec:methodology}
We propose a graph-based  prediction model that exploits Spatial Visibility and Temporal Dynamics. Fig.~\ref{fig:system} presents an overview of our model. 
We divide the  space covered by the entire   point cloud video into multiple equal sized cells. For the example in Fig~\ref{fig:3dcell}, the space is  represented by $5*6*8$ cells and all cells form a grid-like graph.  Neighboring cells in the same frame have strong visibility correlations, which can be exploited by a graph model. Each node in the graph model corresponds to a cell. Each node has its neighboring cells as neighbors in the graph. For example, a cell can have 6 neighbors sharing the same side, or up to 26 neighbors sharing the same corner/edge/side. To simplify the figure, we use a 8-node graph to represent the whole graph in Fig.~\ref{fig:system}. For each node, we have three key features for visibility prediction. 


\subsection{Cell Occupancy Feature}
After partitioning the point cloud video into cells, we can get the number of points in each cell. Since viewer's attention can be driven by the objects in the point cloud video, the point density in a cell will affect the viewer's view interest for it and be an important feature, which we denote as cell occupancy feature. The cell occupancy feature for frame $\tau$ is:
$$
O^\tau = \{ o^{\tau}_i, i\in C \}
$$

\subsection{Cell-based Viewport Feature}
A viewer's viewport can be fully characterized by its 6-DoF coordinates $(x,y,z, yaw,pitch,roll)$. However those coordinates are exogenous to the cell space of the PCV object, and their numeric values have to be pre-processed before feeding to the model,  e.g. the wrap-round from $2\pi$ to $0$ for all the angular features. Instead of using the 6-DoF  coordinates directly, we use the overlap ratio between the current viewport and each cell as the cell-based viewport feature. To simplify the notation, we will use viewport feature in the rest of the paper. To obtain the overlap ratio between the viewport and cell $i$, we randomly generate $N_i$ virutal points in cell $i$, and using the  intrinsic and extrinsic matrices to calculate that. 

More specifically, given a point set $\mathbf{P}$ in world coordinates, transform it to camera coordinates using extrinsic matrix mapping:
\[
\mathbf{P}_{\text{cam}} = \mathbf{E} \cdot \begin{bmatrix} \mathbf{P} \\ 1 \end{bmatrix}
\]
where $\mathbf{E}$ is the extrinsic matrix. And it is determined by the rotation matrix, which is derived from the orientation coordinates $(\psi,\theta, \phi)$.

Then project the points onto the image plane using intrinsic mapping:
\[
\mathbf{P}_{\text{img}} = \mathbf{I} \cdot \mathbf{P}_{\text{cam}}
\]
where $\mathbf{I}$ is the intrinsic matrix. It is determined by the camera's property like focal-length.

After we map all points on the camera's image, we can filter the points in actual FoV based on image dimensions and depth:
\[
\text{In viewport point} = \left\{ \mathbf{P}_{\text{img}} \mid 
\begin{aligned}
  &0 \leq \mathbf{P}_{\text{img}}[0, :] < \text{width}, \\
  &0 \leq \mathbf{P}_{\text{img}}[1, :] < \text{height}, \\
  &d_{\text{near}} < \mathbf{P}_{\text{img}}[2, :] < d_{\text{far}}
\end{aligned}
\right\}
\]
where $d_{\text{near}}$ and $d_{\text{far}}$ are the near and far plane as Fig.~\ref{fig:fov_demo} shows.

By this mapping, we can get the number of points falling into the viewport, as $K_i$. The viewport feature is denoted as:
$$
F^\tau = \left\{ f^{\tau}_i = \frac{K_i}{N_i}, i\in C \right\}
$$

\subsection{Cell Visibility Feature}
As we mentioned earlier, unlike 360-degree video, point cloud video has a unique property that requires consideration of occlusion due to 6DoF (six degrees of freedom). The number of visible points within a cell is crucial for describing occlusion. To estimate the visible points, given the 6DoF viewport coordinates and the point cloud object, we can use the HPR algorithm~\cite{katz2007direct} to determine the number of visible points in a cell. However, applying HPR to high-density point clouds is time-consuming. In~\cite{han2020vivo}, the authors propose a cell-based occlusion estimation method. However, cell-based methods can introduce significant quantization errors. To achieve a better balance between accuracy and computational overhead, we downsample the original point cloud video with a voxel size of 8 to reduce point density and perform HPR at a lower point density to expedite the processm, and then estimate the number of visible points by upsampling. Although downsampling reduces HPR accuracy, it remains more precise than the cell-based occlusion estimation and shows minimal difference from the original resolution. We achieve real-time cell feature calculation at approximately 45 frames per second.

\begin{figure}[htp]
    \centering
    \includegraphics[scale=0.4]{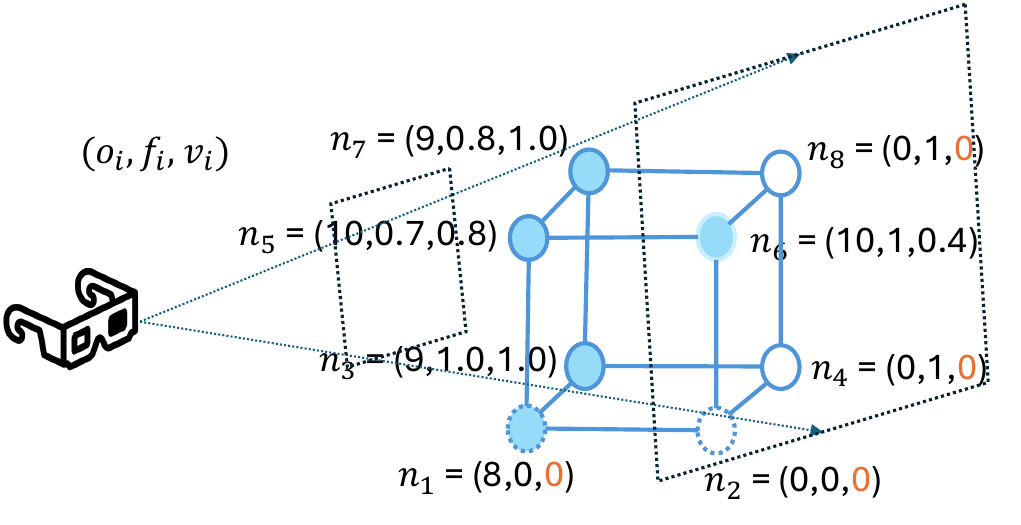}
    \caption{In this illustration, given view's 6DoF and the point cloud frame in 3D cell, we can get the occupancy feature, viewport feature and visibility feature, as $o_i, f_i, v_i$. We have 8 node in total, and 5 of the node with color is occupied by points and has different number of points. $n_6$ has 10 points in total but can be occluded by other points and visibility is 0.4. For other node without points, the visibility feature is set as 0.}
    \label{fig:feature_example_demo}
\end{figure}

After removing the hidden points, we apply intrinsic and extrinsic matrix mapping to the remaining points to exclude those outside the field of view (FoV). The final cell visibility feature is then:
$$
V^\tau = \{ \frac{v^{\tau}_i}{N_i}, i\in C \}. 
$$
 A further illustration for these three features is shown in Fig.~\ref{fig:feature_example_demo}.

\subsection{Other Features}
Other features $E^\tau=\{ E_i^\tau, i \in C^\tau \}  $ for cell $i$  include the cell center coordinates and the distance from the cell center to the viewer's viewpoint. These features help the model account for changes in the viewer's position over time, similar to the way $x,y,z$ coordinates are used in trajectory-based methods. It is important to note that the cell/node index remains fixed throughout the duration of the point cloud video.

\subsection{TransGraph and GRU Model}
We utilize a Transformer-based graph network ~\cite{shi2020masked} to build our graph model. The Transformer-based approach employs attention mechanisms to capture more dynamic relationships between neighbors. Given the hidden state $S_{\tau}$ at time $\tau$, we first concatenate $S_{\tau}$ with the node feature $G_{\tau}$ to form new feature $H_{\tau} = \{h_i^{\tau}, i \in C \}= S_{\tau} \oplus G_{\tau} $ for all cells.  We then update this feature using a weighted average of transformed features at its neigbhoring nodes, described by the following equations:
\begin{equation}
    q_{c,i}^{\tau} = W_{c,q} h_i^{\tau} + b_{c,q}
\end{equation}

\begin{equation}
    k_{c,j}^{\tau} = W_{c,k} h_j^{\tau} + b_{c,k}
\end{equation}

\begin{equation}
    \alpha_{c,ij}^{\tau} = \frac{\langle q_{c,i}^{\tau}, k_{c,j}^{\tau} \rangle}{\sum_{u \in \mathcal{N}(i)} \langle q_{c,i}^{\tau}, k_{c,u}^{\tau} \rangle}
\end{equation}

\begin{equation}
    v_{c,j}^{\tau} = W_{c,v} h_j^{\tau} + b_{c,v}
\end{equation}

\begin{equation}
    \hat{h}_i^{\tau} =  \sum_{j \in \mathcal{N}(i)} \alpha_{c,ij}^{\tau}  v_{c,j}^{\tau}
\end{equation}
For the graph model:
\begin{equation}
\hat{H}_{\tau} = TransGraph(H_{\tau})
\end{equation}
This is a single-layer graph model update; multiple layers will be updated sequentially based on the output of the previous layer. For the GRU model, similar to the approach in ~\cite{he2023graphgru}, we employ a bi-directional GRU to capture temporal patterns. At each time step, the original hidden state $s_i^{\tau}$ at node $i$ is updated to  $\hat{h}_i^{\tau}$ using the graph operation described above, which propagates information from its neighbors in the graph. This updated hidden state will then serve as the hidden state for the next time step in the GRU model in both directions:

\begin{equation}
 S_{t+1} = GRU_f(\hat{H}_{t},G_t)   
\end{equation}

\begin{equation}
S'_{t-1} = GRU_r(\hat{H}'_t,G_t)
\end{equation}
At the end of the GRU modal, a MLP module will take the combined hidden states $S_{h+1}$ and $S_{0}^{'}$ to predict the cell visibility or the cell-based viewport feature.  

Since in a streaming system, when the system decides  to send frame $P^{h+f}$ after encoding it, we can use occupancy feature at time $h+f$, $O^{h+f}$, as the mask on the output of MLP module to get the final prediction $\hat{Y}^{h+f}$. 
$\hat{Y}^{h+f} = M^{h+f} \hat{Z}^{h+f}$
where \( M_{h+f} \) is an indicator variable defined as:
\[
m_i = 
\begin{cases} 
1 & \text{if $o_i \geq$ 0 } \\
0 & \text{otherwise}
\end{cases}
\]
where $Z^{h+f}$ is the output of MLP modal and $M^t = \{m^t_i, i \in C \}$.
A binary mask is applied here because if a cell has no points, there is no need to transmit that cell. Conversely, we do not rely on the number of points within a cell, as having more points does not necessarily equate to higher visibility, which ultimately depends on the viewer's interest.

Combining all the elements, our model is capable of generating the cell feature in real time using the historical viewport trajectory and point cloud video data. Subsequently, the graph model captures the local patterns of cell visibility while the GRU model captures the features that vary over time. The model's output may be cell visibility based on the streaming system or cell-based viewport feature, and the prediction accuracy will be assessed in the following section.

\section{Evaluation}

\subsection{Dataset}
We use a public point cloud video dataset from 8i~\cite{d20178i}, which has four videos, \texttt{longdress}, \texttt{loot}, \texttt{redandblack} and \texttt{soldier}, and a 6DoF viewing navigation trajectory dataset~\cite{subramanyam2020user}. The trajectory data were collected from 26 users watching these four videos at 30 fps. The total watching length is around 40k frames over all the videos. We use the first three videos' trajectory (\texttt{longdress}, \texttt{loot}, \texttt{redandblack}) to train the model and the last one's trajectory (\texttt{soldier}) for testing (the first half) and validation (the second half).

\subsection{Implementation Details}
We pre-process the point cloud videos and generate the cell features on Apple M1 Pro Chip, and train the graph model on NVIDIA A100 GPU. The graph model is based on Pytorch 2.3.0 and CUDA 12.2. Our learning rate is 3e-4, feature dimension is 128. 
The history length is 90 frames (3 seconds) for all the methods except for LR (which is 1 second to have a better performance). We need 22G GPU memory to train the model when the batch size is 32. We set the total epoch number as 30, with early stopping on validation data when validation loss does not reduce for 5 epochs. We partition the whole space into 5*6*8=240 cells. The point cloud video has around 1.8m in height (Y axis) in physical world and the cubic cell size is around 0.2m along all three  dimensions. The camera image resolution is (1920*1080). The intrinsic matrix of camera is
$\begin{bmatrix}
f_x & 0 & c_x \\
0 & f_y & c_y \\
0 & 0 & 1
\end{bmatrix}$, where $f_x=f_y=525$, $c_x = 1920/2$, $c_y = 1080/2$.


\subsection{Baseline Methods and Evaluation Metrics}

We compare our proposed model against several trajectory-based baseline models to evaluate its performance in predicting cell visibility. The trajectory-based models first predict the future viewport 6DoF coordinates \((\hat{x}, \hat{y}, \hat{z}, \hat{\psi}, \hat{\theta}, \hat{\phi})\) based on the historical viewport trajectory. These predicted coordinates are then used to calculate cell visibility and the cell-based viewport features defined in the Methodology section. In contrast, our model directly predicts cell visibility and the cell-based viewport features. 
We evaluate the performance of our model and the baselines by comparing the average Mean Squared Error (MSE) loss across all cells. Additionally, we use the \(R^2\) score to assess the variation and confidence of the predictions. The \(R^2\) score, also known as the coefficient of determination, measures the proportion of variance in the dependent variable that is predictable from the independent variables. An \(R^2\) score of 1 indicates perfect prediction, while a score of 0 indicates that the model does not explain any of the variability in the data. For trajectory-based methods, the angles \(\psi\), \(\theta\), and \(\phi\) can be challenging to handle because the value wrap-around, e.g. from \(2\pi\) to \(0\). There are different approaches to address this issue. In our approach, we convert the orientation coordinates to the sine and cosine domains, using two coordinates to represent each angle coordinate. After prediction, we convert these coordinates back to the original angle value using the arctangent function. The baselines are as follows:
\begin{itemize}
\item Linear Regression (LR): We use LR to predict all coordinate individually.  A linear model predicts each of the 6DoF coordinate using a linear combination of the values on the same coordinate over a  history window. This model serves as a fundamental baseline for comparison, and shows more FoV prediction accuracy than MLP sometimes~\cite{han2020vivo}. 
\item Truncated linear regression (TLR): This approach uses the last monotonically increasing or decreasing part of the history window to linearly extrapolate the future value. It has shown a great performance in sequence prediction, in particular  
short-term FoV prediction~\cite{li2019very}. Similar to LR, we use TLR to predict all coordinates individually. 
\item Mutli-task Multilayer Perceptron (M-MLP) model in ~\cite{hou2020motion}: A feedforward neural network with one or more hidden layers is used for prediction. The MLP can capture non-linear relationships between input features and the future FoV, providing a more complex baseline than LR. 
As~\cite{hou2020motion} shows, predicting all the coordinates together using all the coordinates in the history window can improve the performace. Therefore we use a MLP to predict all coordinates together. Similar to the MLP model in ~\cite{han2020vivo, hou2020motion,liu2023cav3}, we use a MLP model comprising two hidden layers, with 60 neurons in each layer, following a fully connected architecture. Activation function is `ReLu'.
\item Mutli-task LSTM(M-LSTM): The LSTM, a type of recurrent neural network (RNN), is well-suited for sequence prediction, learning long-term dependencies in time series data. Similar to ~\cite{hou2020motion}, we use a two-layer LSTM model with 60 neurons per layer to predict future FoV coordinates based on historical data, predicting all coordinates simultaneously
\end{itemize}


\subsubsection{Visibility Prediction}
\begin{figure}[htp]
    \centering
    \includegraphics[scale=0.3]{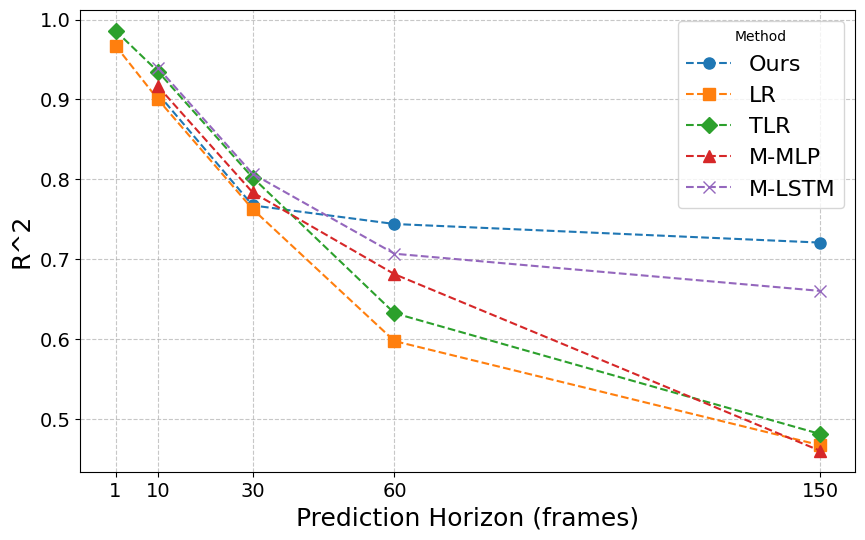}
    \caption{$R^2$ Score of Cell Visibility Prediction at Different Prediction Horizons.}
    \label{fig:visibility}
\end{figure}

In Fig.~\ref{fig:visibility}, we evaluate our model across different prediction horizons. As expected, simpler models like LR and TLR perform well for short horizons. However, our model significantly improves prediction accuracy for horizons extending beyond 1 second.

\begin{table}[h]
\centering
\begin{tabular}{|c|c|c|c|c|}
\hline
\textbf{Method} & \textbf{333ms} & \textbf{1000ms} & \textbf{2000ms} & \textbf{5000ms} \\ \hline
\textbf{LR} & 0.0043 & 0.0102 & 0.0173 & 0.0229 \\ \hline
\textbf{TLR} & 0.0028 & 0.0085 & 0.0158 & 0.0223 \\ \hline
\textbf{M-MLP} & 0.0036 & 0.0093 & 0.0137 & 0.0232 \\ \hline
\textbf{M-LSTM} & \textbf{0.0026} & \textbf{0.0083} &0.0126 & 0.0146 \\ \hline
\textbf{Ours} & 0.004 & 0.0100 & \textbf{0.0110} & \textbf{0.0120} \\ \hline
\end{tabular}
\caption{MSE of Visibility Prediction by Different Methods at Different Prediction Horizons}
\label{tab:mse_loss_methods}
\end{table}
In Table.~\ref{tab:mse_loss_methods}, we report the MSE losses for different methods across various prediction horizons. For short-term predictions (less than 1000 ms), our model maintains a relatively consistent cell visibility prediction loss. More importantly, for long-term cell visibility predictions, our model reduces the MSE loss by up to 20\% compared to all state-of-the-art methods. This improvement is significant for on-demand point cloud video streaming with target buffer length around 5 second. Our model effectively addresses the error amplification issue of trajectory-based methods, and captures the temporal and spatial patterns in viewer's attention and cell visibility.

\subsubsection{Cell-based Viewport Prediction}
\begin{figure}[ht]
    \centering
    \includegraphics[scale=0.3]{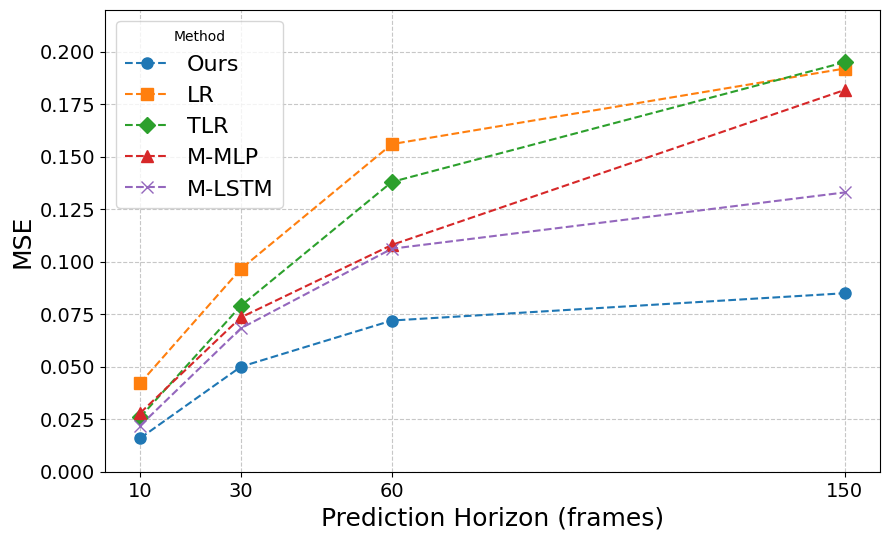}
    \caption{Viewport prediction loss with different prediction horizon, from 10 frames(333ms) to 150 frames(5000ms).}
    \label{fig:fov-viewport prediction}
\end{figure}

\begin{figure}[ht]
    \centering
    \includegraphics[width=0.9\linewidth]{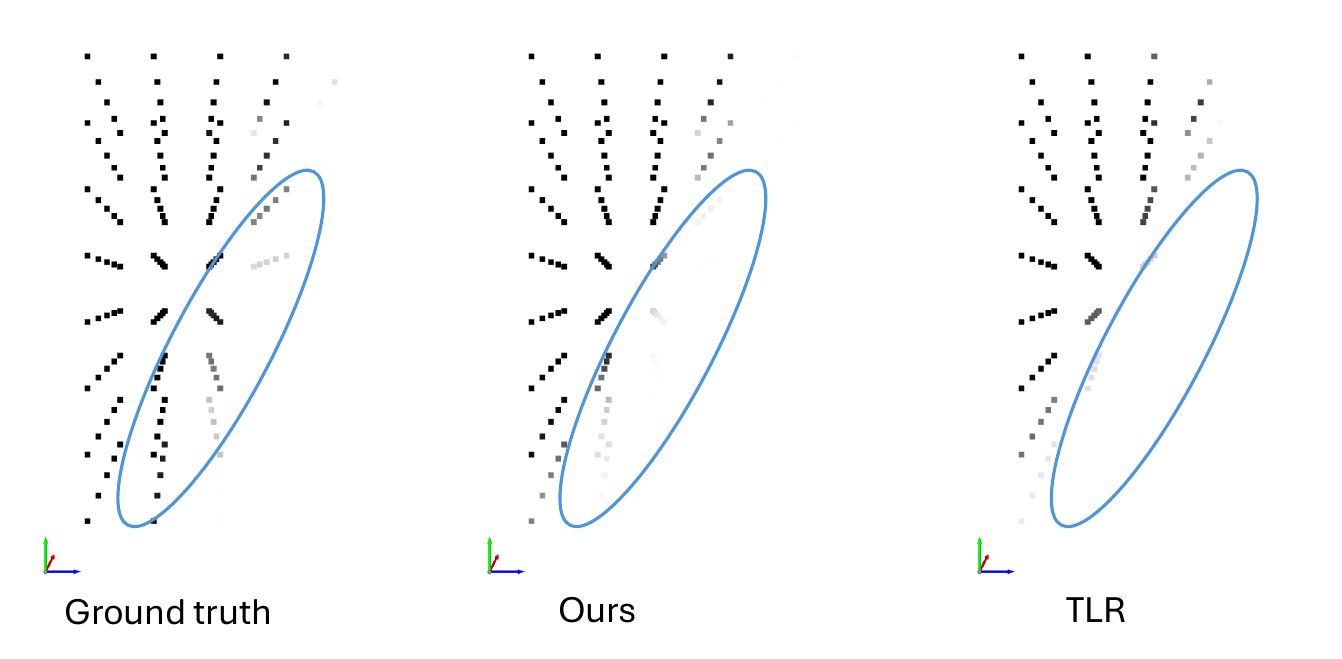}
    \caption{A visualization of predicted viewport in 3D space. Each dot represent one cell in our grid-like graph. The heavier color means higher visibility for this cell. Comparison of Ground Truth, Our Prediction, and TLR. The circled part highlights the marginal prediction difference among different methods.}
    \label{fig:viewportcasestudy}
\end{figure}

Our model can also be used to directly predict the viewport by forecasting the cell-based viewport feature defined in the Methodology Section. As illurated in Figure~\ref{fig:system}, by setting $\hat{Y}_{h+f} = \hat{F}_{h+f}$, our model can  predict the overlap ratios between the viewport and all the cells (essentially predicting the viewport). We compare our model with all the baselines for viewport prediction across different horizons in Fig.~\ref{fig:fov-viewport prediction}. Our model consistently outperforms all the state-of-the-art baselines across all prediction horizons, demonstrating the superiority of our spatial perception method over the traditional trajectory-based approaches. The MSE loss for our model is 0.06, while the TLR loss is 0.13, corresponding to the average MSE loss for a 2 second (or 60 frames) prediction horizon shown in Fig.~\ref{fig:fov-viewport prediction}. The performance gaps widen at the longer prediction horizon of 5 second (or 150 frames). Additionally, a case study visualizing the viewport prediction is presented in Fig.~\ref{fig:viewportcasestudy}. Our model's predictions are much closer to the ground truth, especially in the marginal areas, which are typically challenging for traditional trajectory-based methods to predict accurately.

\subsection{Cell in FoV Correlation}
\label{subsec:correlation}
\begin{figure}[ht]
    \centering
    \includegraphics[width=0.9\linewidth]{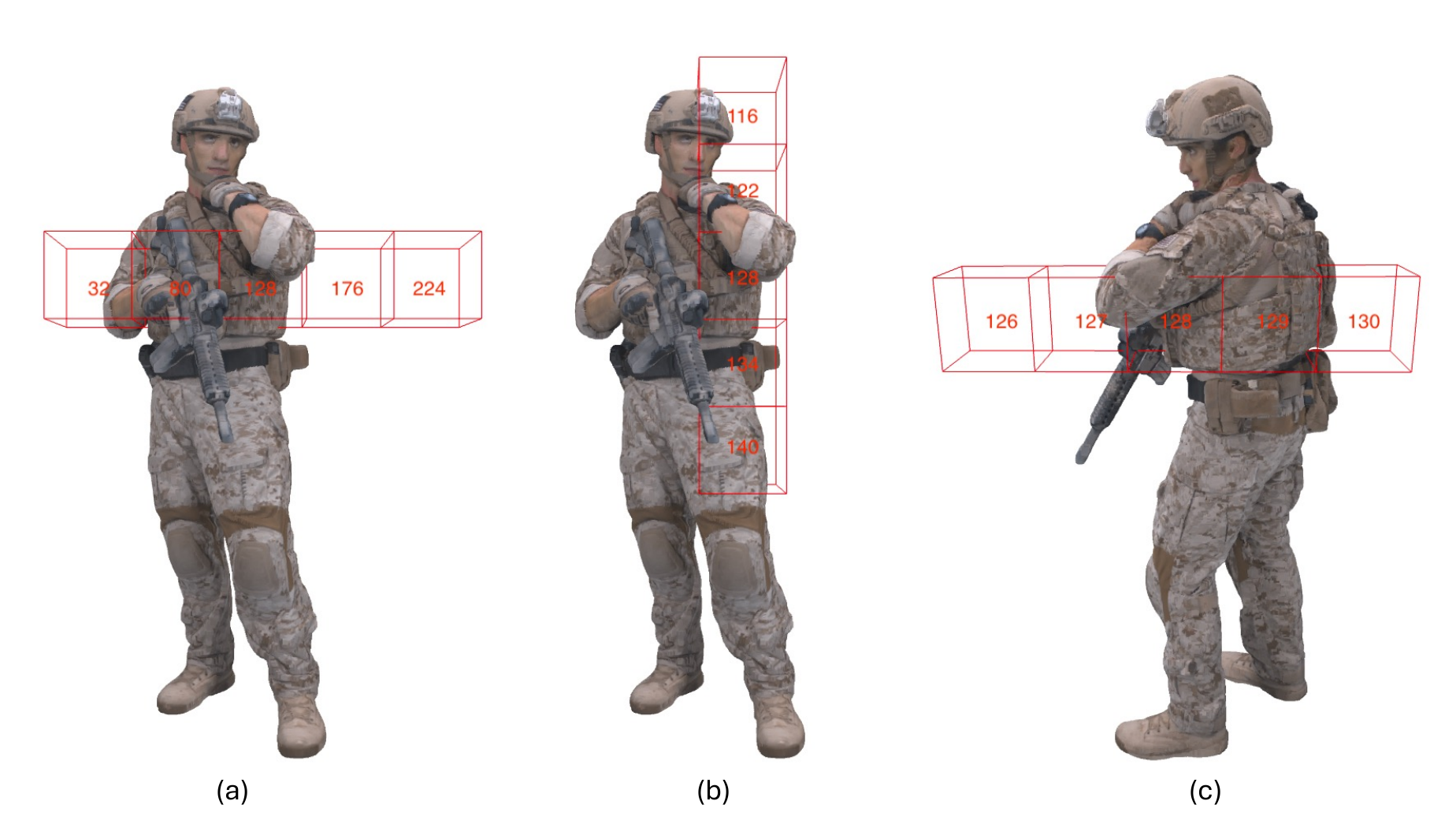}
    \caption{The relative positions for the example cell. The figure shows the position of cell 128 and its neighboring cells in different directions. (a) illustrates cell 128 and its neighbors along the x-axis. (b) depicts cell 128 and its neighbors along the y-axis. (c) displays cell 128 and its neighbors along the z-axis.}
    \label{fig:coorelation}
\end{figure}

\begin{figure}[ht]
    \centering
    \includegraphics[width=1\linewidth]{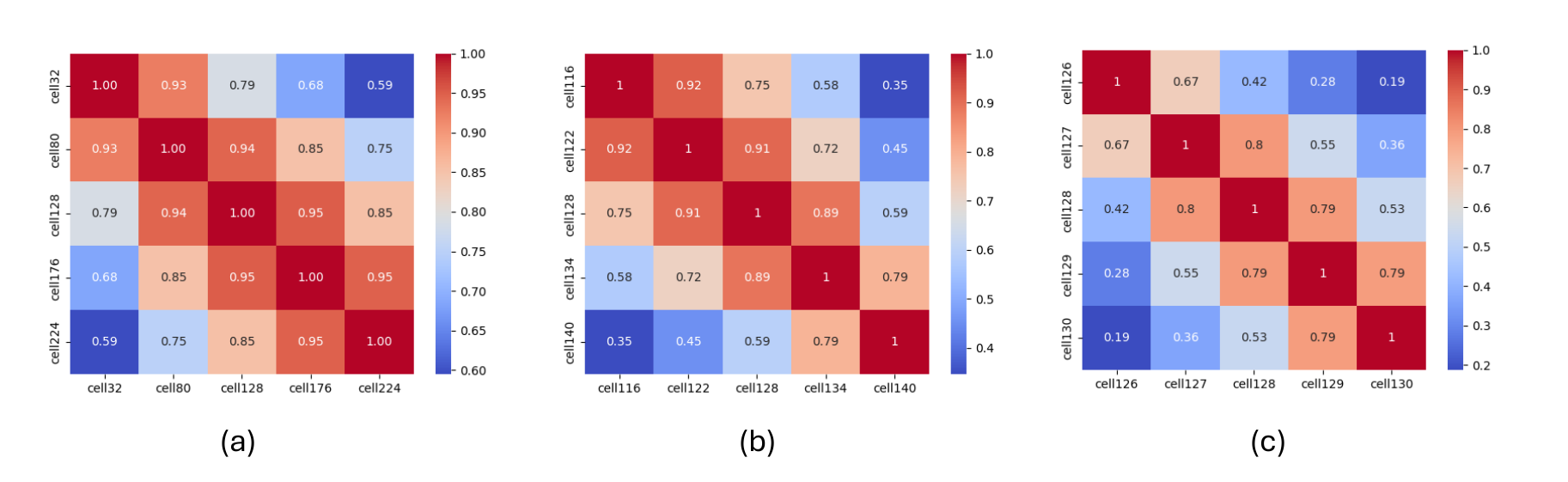}
    \caption{The correlation of the viewport feature indicates the fraction of the cell's volume that is within the FoV. (a) shows the correlation among five example cells along the x-axis. (b) and (c) display the correlation among the example cells along the y-axis and z-axis, respectively. }
    \label{fig:correlation_heapmap}
\end{figure}

As shown in Fig.~\ref{fig:coorelation} and Fig.~\ref{fig:correlation_heapmap}, the viewport feature exhibits a strong correlation with neighboring cells. We calculated this correlation using FoV trajectories from 26 users for the \texttt{soldier}  video. Neighbors that are closer together show stronger correlations compared to those that are farther apart. This strong correlation offers valuable insights for predicting whether a cell will enter or exit the user's FoV based on the state of its neighboring cells. By leveraging transformer-based graph neural networks, our model can capture this dynamic pattern and enhance its ability to predict visibility.

\vspace{-0.3in}

\vspace{-0.3in}

\section{Conclusion}
In this paper, we introduce a novel spatial-based FoV prediction approach designed to predict long-term cell visibility for PCV. Our method leverages both spatial and temporal dynamics of PCV objects and viewers, outperforming existing state-of-the-art methods in terms of prediction accuracy and robustness. By integrating Transformer-based GNNs and graph attention networks, our model efficiently captures complex relationships between neighboring cells with a single graph layer. This approach overcomes the limitations of trajectory-based FoV prediction by incorporating the full spatial context, resulting in more accurate and stable predictions. Our spatial-based FoV prediction model presents a promising solution for long-term 6-DoF FoV prediction, immersive video streaming, and 3D rendering. We will make the code available to support further research and development in this area.

\section{Acknowledgement}
This project was partially supported by USA NSF under award number CNS-2312839.


\begin{thebibliography}{19}
\providecommand{\natexlab}[1]{#1}

\bibitem[{Ban et~al.(2018)Ban, Xie, Xu, Zhang, Guo, and Wang}]{ban2018cub360}
Ban, Y.; Xie, L.; Xu, Z.; Zhang, X.; Guo, Z.; and Wang, Y. 2018.
\newblock Cub360: Exploiting cross-users behaviors for viewport prediction in 360 video adaptive streaming.
\newblock In \emph{2018 IEEE International Conference on Multimedia and Expo (ICME)}, 1--6. IEEE.

\bibitem[{Dong et~al.(2023)Dong, Shen, Xie, Li, Zuo, and Zhang}]{longterm360Dong}
Dong, P.; Shen, R.; Xie, X.; Li, Y.; Zuo, Y.; and Zhang, L. 2023.
\newblock Predicting Long-Term Field of View in 360-Degree Video Streaming.
\newblock \emph{IEEE Network}, 37(3): 26--33.

\bibitem[{d’Eon et~al.(2017)d’Eon, Harrison, Myers, and Chou}]{d20178i}
d’Eon, E.; Harrison, B.; Myers, T.; and Chou, P.~A. 2017.
\newblock 8i voxelized full bodies-a voxelized point cloud dataset.
\newblock \emph{ISO/IEC JTC1/SC29 Joint WG11/WG1 (MPEG/JPEG) input document WG11M40059/WG1M74006}, 7(8): 11.

\bibitem[{Fan et~al.(2017)Fan, Lee, Lo, Huang, Chen, and Hsu}]{fan2017fixation}
Fan, C.-L.; Lee, J.; Lo, W.-C.; Huang, C.-Y.; Chen, K.-T.; and Hsu, C.-H. 2017.
\newblock Fixation prediction for 360 video streaming in head-mounted virtual reality.
\newblock In \emph{Proceedings of the 27th workshop on network and operating systems support for digital audio and video}, 67--72.

\bibitem[{Han, Liu, and Qian(2020)}]{han2020vivo}
Han, B.; Liu, Y.; and Qian, F. 2020.
\newblock ViVo: Visibility-aware mobile volumetric video streaming.
\newblock In \emph{Proceedings of the 26th annual international conference on mobile computing and networking}, 1--13.

\bibitem[{He, Su, and Ye(2023)}]{he2023graphgru}
He, H.; Su, L.; and Ye, K. 2023.
\newblock Graphgru: A graph neural network model for resource prediction in microservice cluster.
\newblock In \emph{2022 IEEE 28th International Conference on Parallel and Distributed Systems (ICPADS)}, 499--506. IEEE.

\bibitem[{Hou and Dey(2020)}]{hou2020motion}
Hou, X.; and Dey, S. 2020.
\newblock Motion Prediction and Pre-Rendering at the Edge to Enable Ultra-Low Latency Mobile 6DoF Experiences. IEEE Open Journal of the Communications Society 1 (2020), 1674--1690.

\bibitem[{Katz, Tal, and Basri(2007)}]{katz2007direct}
Katz, S.; Tal, A.; and Basri, R. 2007.
\newblock Direct visibility of point sets.
\newblock In \emph{ACM SIGGRAPH 2007 papers}, 24--es.

\bibitem[{Lee et~al.(2020)Lee, Yi, Lee, Choi, and Kim}]{groot2020}
Lee, K.; Yi, J.; Lee, Y.; Choi, S.; and Kim, Y.~M. 2020.
\newblock GROOT: A Real-Time Streaming System of High-Fidelity Volumetric Videos.
\newblock In \emph{Proceedings of the 26th Annual International Conference on Mobile Computing and Networking}, MobiCom '20. New York, NY, USA: Association for Computing Machinery.
\newblock ISBN 9781450370851.

\bibitem[{Li et~al.(2023)Li, Wang, Zong, Cao, and Liu}]{li2023predictive}
Li, C.; Wang, X.; Zong, T.; Cao, H.; and Liu, Y. 2023.
\newblock Predictive edge caching through deep mining of sequential patterns in user content retrievals.
\newblock \emph{Computer Networks}, 233: 109866.

\bibitem[{Li et~al.(2019)Li, Zhang, Liu, and Wang}]{li2019very}
Li, C.; Zhang, W.; Liu, Y.; and Wang, Y. 2019.
\newblock Very long term field of view prediction for 360-degree video streaming.
\newblock In \emph{2019 IEEE Conference on Multimedia Information Processing and Retrieval (MIPR)}, 297--302. IEEE.

\bibitem[{Liu et~al.(2023)Liu, Zhu, Wang, Jin, Zhang, Xu, and Cui}]{liu2023cav3}
Liu, J.; Zhu, B.; Wang, F.; Jin, Y.; Zhang, W.; Xu, Z.; and Cui, S. 2023.
\newblock Cav3: Cache-assisted viewport adaptive volumetric video streaming.
\newblock In \emph{2023 IEEE Conference Virtual Reality and 3D User Interfaces (VR)}, 173--183. IEEE.

\bibitem[{Qian et~al.(2016)Qian, Ji, Han, and Gopalakrishnan}]{qian2016optimizing}
Qian, F.; Ji, L.; Han, B.; and Gopalakrishnan, V. 2016.
\newblock Optimizing 360 video delivery over cellular networks.
\newblock In \emph{Proceedings of the 5th Workshop on All Things Cellular: Operations, Applications and Challenges}, 1--6.

\bibitem[{Shi et~al.(2020)Shi, Huang, Feng, Zhong, Wang, and Sun}]{shi2020masked}
Shi, Y.; Huang, Z.; Feng, S.; Zhong, H.; Wang, W.; and Sun, Y. 2020.
\newblock Masked label prediction: Unified message passing model for semi-supervised classification.
\newblock \emph{arXiv preprint arXiv:2009.03509}.

\bibitem[{Subramanyam et~al.(2020)Subramanyam, Viola, Hanjalic, and Cesar}]{subramanyam2020user}
Subramanyam, S.; Viola, I.; Hanjalic, A.; and Cesar, P. 2020.
\newblock User centered adaptive streaming of dynamic point clouds with low complexity tiling.
\newblock In \emph{Proceedings of the 28th ACM international conference on multimedia}, 3669--3677.

\bibitem[{Sun et~al.(2019)Sun, Duanmu, Liu, Wang, Ye, Shi, and Dai}]{sun2019two}
Sun, L.; Duanmu, F.; Liu, Y.; Wang, Y.; Ye, Y.; Shi, H.; and Dai, D. 2019.
\newblock A two-tier system for on-demand streaming of 360 degree video over dynamic networks.
\newblock \emph{IEEE Journal on Emerging and Selected Topics in Circuits and Systems}, 9(1): 43--57.

\bibitem[{Wang et~al.(2021)Wang, Li, Dai, Zou, and Xiong}]{wang2021qoe}
Wang, L.; Li, C.; Dai, W.; Zou, J.; and Xiong, H. 2021.
\newblock QoE-driven and tile-based adaptive streaming for point clouds.
\newblock In \emph{ICASSP 2021-2021 IEEE International Conference on Acoustics, Speech and Signal Processing (ICASSP)}, 1930--1934. IEEE.

\bibitem[{Wu et~al.(2020)Wu, Zhang, Wang, and Sun}]{wu2020spherical}
Wu, C.; Zhang, R.; Wang, Z.; and Sun, L. 2020.
\newblock A spherical convolution approach for learning long term viewport prediction in 360 immersive video.
\newblock In \emph{Proceedings of the AAAI Conference on Artificial Intelligence}, volume~34, 14003--14040.

\bibitem[{Zong et~al.(2023)Zong, Mao, Li, Liu, and Wang}]{zong2023progressive}
Zong, T.; Mao, Y.; Li, C.; Liu, Y.; and Wang, Y. 2023.
\newblock Progressive Frame Patching for FoV-based Point Cloud Video Streaming.
\newblock \emph{arXiv preprint arXiv:2303.08336}.

\end{thebibliography}
\end{document}